\documentclass[letterpaper, 10 pt, conference]{ieeeconf}  %

\IEEEoverridecommandlockouts                              %

\usepackage{graphicx}
\usepackage{amsmath}
\usepackage{cite}
\usepackage[backref, colorlinks, linkcolor={blue!50!black}, citecolor={blue!50!black}, urlcolor={blue!80!black}]{hyperref}
\usepackage[capitalise]{cleveref}
\usepackage{bm}
\usepackage{bbm}
\usepackage{amssymb}
\usepackage{amsmath}
\usepackage{booktabs}
\usepackage{tabularx}
\usepackage{overpic}
\usepackage{contour}
\usepackage{colortbl}
\usepackage{siunitx}
\usepackage{flushend}
\usepackage[table]{xcolor}
\usepackage{url}
\usepackage{breakurl}
\definecolor{pastelBlue}{rgb}{0.68, 0.85, 0.9}
\definecolor{darkBlue}{rgb}{0.34, 0.425, 0.45}

\usepackage{tikz}
\usetikzlibrary{arrows.meta, shadows, fadings,shapes.arrows}
\tikzset{My Arrow Style/.style={single arrow, fill=pastelBlue, anchor=base, align=center,text width=3cm, text height=0.5cm, text=darkBlue}}

\contourlength{0.05em} %
\contournumber{50} %

\newcommand{\etal}{\textit{et al}. }

\newcommand{\secref}[1]{Sec.~\ref{#1}}

\newcommand{\figref}[1]{Fig.~\ref{#1}}
\newcommand{\tabref}[1]{Table~\ref{#1}}

\newcommand{\RNum}[1]{\uppercase\expandafter{\romannumeral #1\relax}}

\newcommand{\onimagetext}[1]{\fcolorbox{black}{black}{{\textbf{\textcolor{white}{#1}}}}}

\crefformat{equation}{(#2#1#3)}

\title{\LARGE \bf
Perceptive Pedipulation with Local Obstacle Avoidance
}

\author{Jonas Stolle, Philip Arm, Mayank Mittal, Marco Hutter%
\thanks{All authors are with ETH Zurich, Robotics Systems Lab; Leonhardstrasse 21, 8092 Zurich, Switzerland. M. Mittal is also with NVIDIA.
        Contact: {\tt\small \{jstolle, parm, mittalma\}@ethz.ch}}%
\thanks{This research was supported by the Swiss National Science Foundation through the National Centre of Competence in Digital Fabrication (NCCR dfab). This project has received funding through ESA contract nos. 4000137333/22/NL/AT and 4000135310/21/NL/PA/pt. This work has been conducted as part of ANYmal Research, a community to advance legged robotics.
}
}

\begin{document}

\maketitle
\thispagestyle{empty}
\pagestyle{empty}

\begin{abstract}
    Pedipulation leverages the feet of legged robots for mobile manipulation, eliminating the need for dedicated robotic arms.
    While previous works have showcased blind and task-specific pedipulation skills, they fail to account for static and dynamic obstacles in the environment.
    To address this limitation, we introduce a reinforcement learning-based approach to train a whole-body obstacle-aware policy that tracks foot position commands while simultaneously avoiding obstacles.
    Despite training the policy in only five different static scenarios in simulation, we show that it generalizes to unknown environments with different numbers and types of obstacles.
    We analyze the performance of our method through a set of simulation experiments and successfully deploy the learned policy on the ANYmal quadruped, demonstrating its capability to follow foot commands while navigating around static and dynamic obstacles.
    Videos of the experiments are available at \href{https://sites.google.com/leggedrobotics.com/perceptive-pedipulation}{sites.google.com/leggedrobotics.com/perceptive-pedipulation}.
\vspace{15pt}
\end{abstract}

\section{Introduction}
\label{sec:introduction}
Robotic manipulation skills have rapidly improved in recent years, powered by learning-based methods and increased computational resources. Legged mobile manipulation, in particular, is becoming increasingly feasible, by combining recent advances in robust quadrupedal locomotion \cite{miki2022learning, rudin2022learningwalkminutesusing} with dedicated manipulator arms \cite{fu2022deep, Mittal_2022}. However, robotic arms increase the mass and complexity of the system and reduce the capacity for additional payloads.

Recent works have explored \emph{pedipulation} -- manipulation using a quadrupedal robot's foot -- to overcome these limitations~\cite{arm2024pedipulate, cheng2023legs, he2024learning}.
These works use reinforcement learning (RL) to track position commands for the pedipulating foot with an emerging whole-body behavior.
They successfully demonstrate skills such as pushing a button, opening a door, and lifting objects using the foot as an end-effector.
However, these methods primarily rely on proprioceptive information and do not consider obstacles in the environment. Thus, the learned behaviors are unable to avoid unwanted collisions.

A potential approach to address this limitation is to develop an obstacle-avoiding high-level controller in a hierarchical architecture, leveraging the existing blind policies for low-level joint control.
However, since the high-level planner can only command the pedipulating foot's position without direct control over the remaining degrees of freedom (DoF), collisions between other parts of the robot and the environment may still occur during command tracking by the blind policy.

\begin{figure}[t]
    \centering
    \vspace{6pt}
    \begin{overpic}[width=\linewidth]{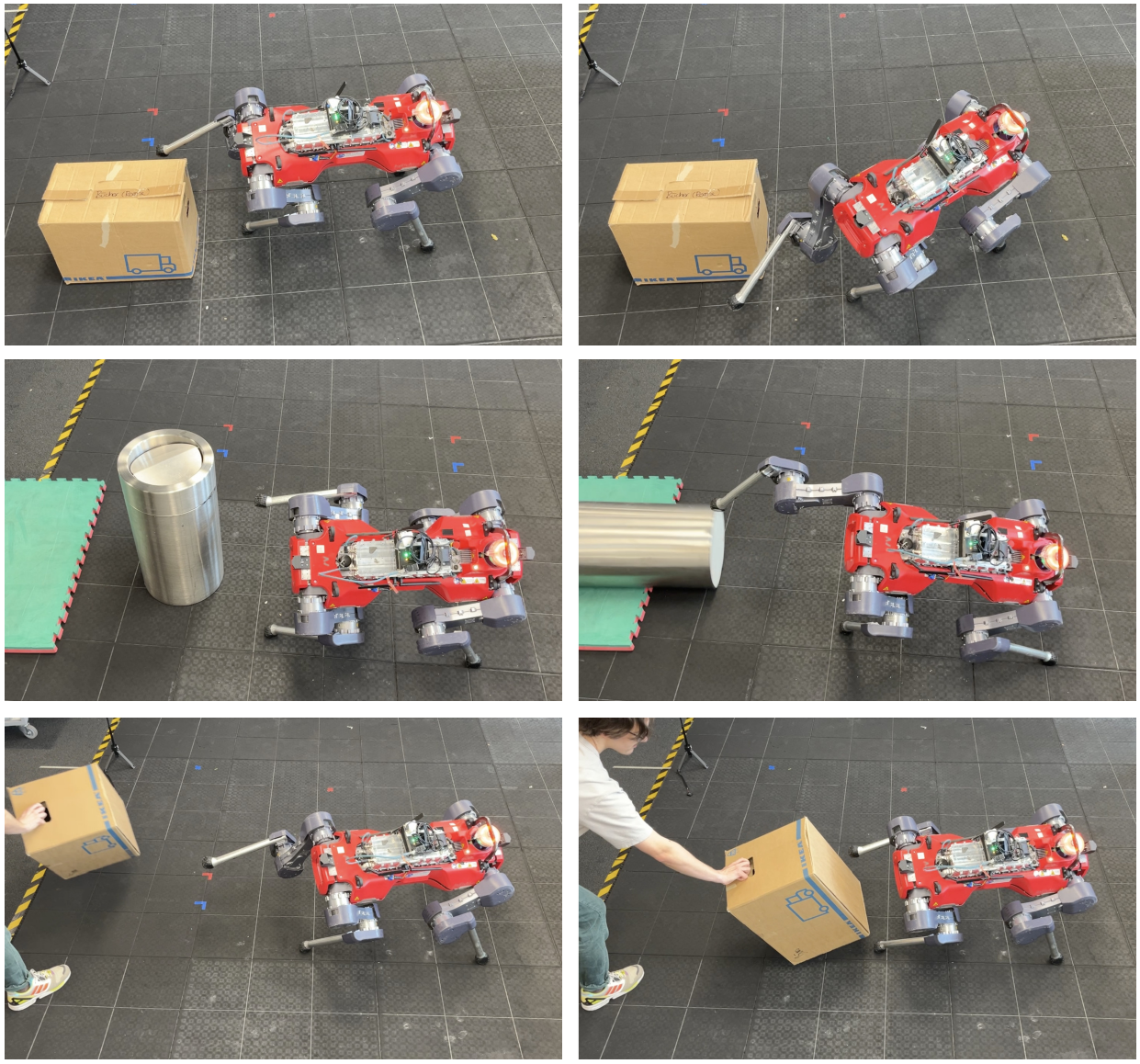}
        \put(2.5,87.0){\onimagetext{A}} %
        \put(2.5,56.0){\onimagetext{B}} %
        \put(2.5,25.0){\onimagetext{C}} %

    \end{overpic}
    \caption{Our perceptive pedipulation controller avoids obstacles with the pedipulating foot of a quadruped robot (A). It can also react to and avoid dynamic obstacles (B). By toggling the contact switch, the robot can push obstacles out of the way on command (C).}
    \label{fig:eyecatcher}
    \vspace{-8pt}
\end{figure}

Thus, instead of a hierarchical design, this work proposes to extend the existing joint-level controllers by integrating perception in an end-to-end fashion. Such a policy retains the simple interface from prior works~\cite{arm2024pedipulate,cheng2023legs,he2024learning}, 
requiring only a foot position command while extending the system's capability to simultaneously perform whole-body obstacle avoidance.
We train this policy entirely in simulation using simple arrangements of cuboidal obstacles and providing appropriate collision rewards to learn the desired collision avoidance behavior. In addition, we introduce a switch to allow or disallow contacts with the pedipulating foot.
With this contact switch, the policy can either avoid obstacles entirely or perform pedipulation tasks while avoiding obstacles with the rest of the body, retaining the capabilities from our previous work~\cite{arm2024pedipulate}.

The key contributions of this work are as follows: 
\begin{itemize}
    \item We design an RL-based perceptive pedipulation controller that tracks foot position commands while avoiding obstacles in the robot's vicinity (\figref{fig:eyecatcher}).
    \item We show that our controller’s obstacle-avoidance behavior generalizes to unseen geometries and effectively avoids dynamic obstacles despite being trained solely on a few scenarios with non-moving obstacles.
    \item We introduce a contact switch for the pedipulating foot, enabling both whole-body obstacle avoidance and pedipulation capabilities while avoiding obstacles with the remaining parts of the robot.
\end{itemize}

\section{Related Work}
\label{sec:related_work}
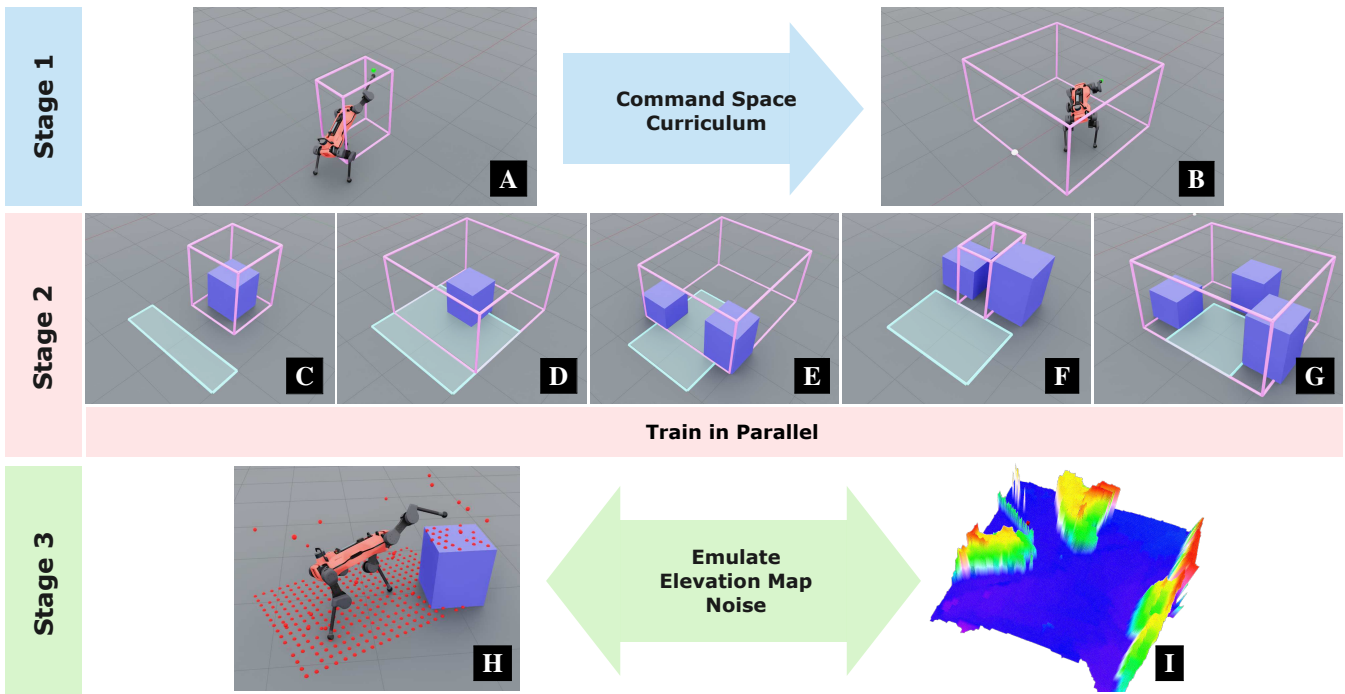
\begin{figure*}[t]
    \centering
    \vspace{8pt}
    \begin{overpic}[width=\textwidth]{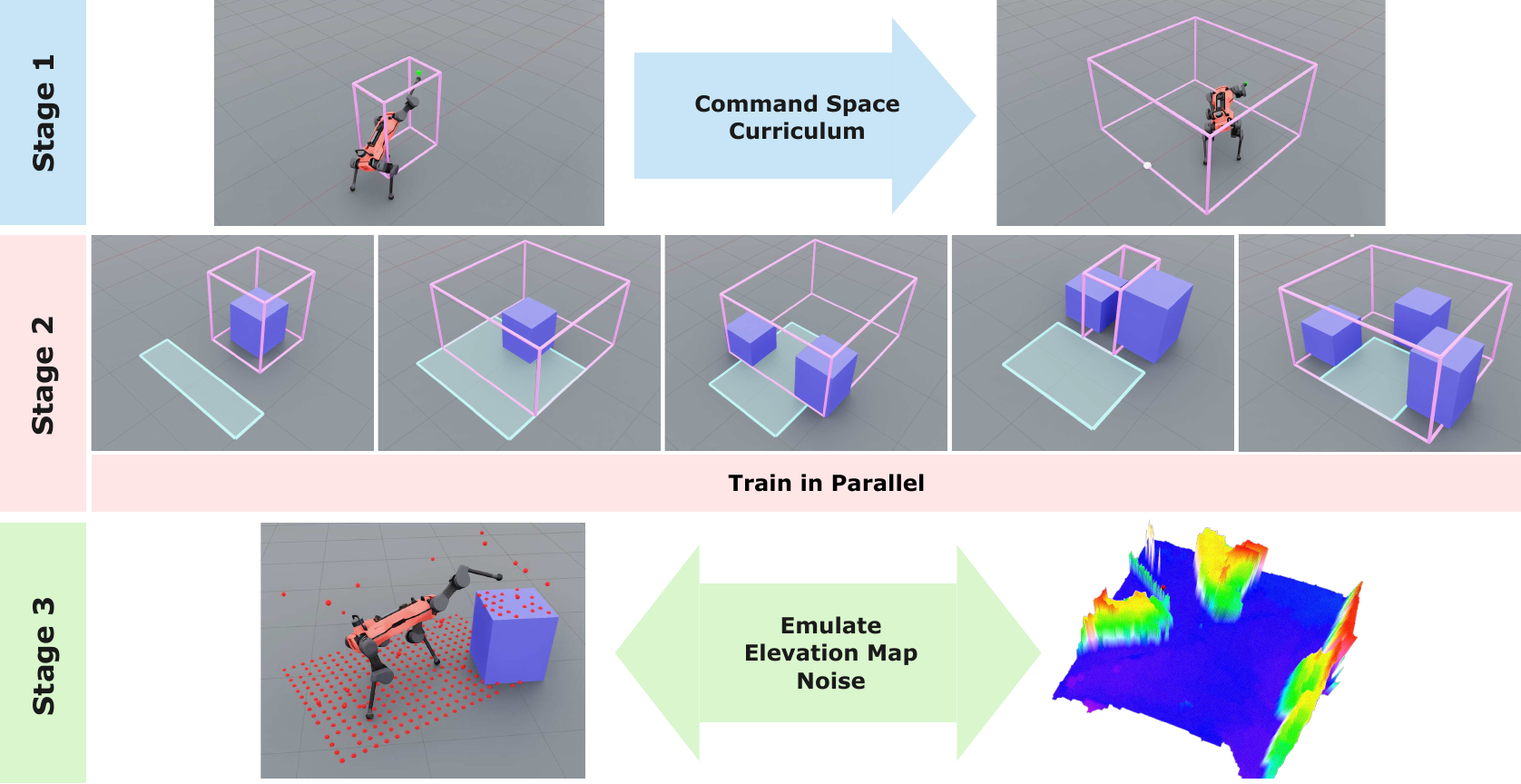}
        \put(36.25,38){\onimagetext{A}}
        \put(87.75,38){\onimagetext{B}}
        \put(21,23.25){\onimagetext{C}}
        \put(40,23.25){\onimagetext{D}}
        \put(59,23.25){\onimagetext{E}}
        \put(77.75,23.25){\onimagetext{F}}
        \put(96.5,23.25){\onimagetext{G}}
        \put(35,1.75){\onimagetext{H}}
        \put(86,1.75){\onimagetext{I}}
    \end{overpic}
    \caption{High-level overview of the training process. The command spaces are denoted in pink, and the spawn spaces are denoted in cyan. In Stage 1, we learn obstacle-free pedipulation following \cite{arm2024pedipulate}. Stage 2 introduces obstacles arranged in five scenarios, where the policy learns obstacle avoidance. Stage 3 increases the noise on the perceptive observations (H) to enable dealing with the noise and artifacts in the mapping pipeline during deployment (I).}
    \label{fig:stages}

\end{figure*}

\subsection*{Pedipulation}
While dedicated robotic arms can be added to quadrupeds for manipulation~\cite{chen_2022, Mittal_2022, chiu2022collisionfree, fu2022deep}, they increase the mass and complexity of the system and reduce the capacity for additional payloads.
Quadrupeds without a separate manipulation arm can still exhibit manipulation capabilities, for instance, object reorientation~\cite{shi2020circus} and whole-body object repositioning~\cite{jeon2023learning}. 

For general and precise manipulation tasks, recent works have looked into pedipulation - the manipulation of objects using a robot’s foot~\cite{cheng2023legs,he2024learning,arm2024pedipulate, lin2024locoman}. These works specify a desired position for the pedipulating foot and control it with learning-based methods in a whole-body fashion. This approach provides a simple control interface for a human operator or a high-level planner to perform various tasks. However, the learned pedipulation policy is limited to obstacle-free environments, as it relies only on proprioceptive information. %
In this work, we aim to account for obstacles in the robot's workspace by incorporating perceptive observations.

Similar to the prior works, we also choose RL to learn pedipulation, as we want to avoid specifying gait patterns that are typically required for online model-based methods~\cite{bjelonic2021mpcwheeled, villarreal2020mpclocomotion}. While model-based methods can solve locomotion tasks successfully, in the context of pedipulation, where the robot has to stand and locomote on three legs, providing the gait patterns can be difficult and unintuitive. Learning-based methods can help overcome this limitation, as the foot positions are unconstrained, and adaptive behaviors can emerge during training~\cite{rudin2022advanced}.

\subsection*{Obstacle Avoidance}

While many classical approaches exist for obstacle avoidance, they are hard to apply to systems with high degrees of freedom (DoF). Grid-based methods suffer from exploding computational costs for increasing DoFs, while sampling-based methods cannot guarantee solutions~\cite{petrović2018motion}.
Model-based approaches can find collision-free whole-body trajectories for high DoF systems~\cite{chiu2022collisionfree} but are hard to apply to our problem because we want to avoid pre-specifying the gait sequence. 

Recently, researchers have also been investigating learning-based methods for obstacle avoidance~\cite{app13148174}.
Miki~\etal~\cite{miki2022learning} use perception to enable collision-free locomotion in confined spaces by training a high-level policy that commands the base pose and using a low-level policy to track this command. As motivated in~\secref{sec:introduction}, this decoupling is limiting in the case of pedipulation.
Honerkamp~\etal~\cite{honerkamp2022learning} learn obstacle-avoiding manipulation skills for a wheeled-mobile base with an attached manipulator arm. In their work, the learned policy outputs the end-effector and base velocity commands, which are converted to the desired joint positions through inverse kinematics (IK).
While it is possible to follow a similar approach that separates the system into a three-legged mobile base and the pedipulating leg, using IK reduces the whole-body reachability, as shown in~\cite{arm2024pedipulate}.

\section{Method}
\label{sec:task_description}

This work builds on our previous work~\cite{arm2024pedipulate}, where we used RL to train a foot position tracking policy and showed successful sim-to-real transfer. We extend this work by incorporating perception and obstacles during training to enable automatic collision avoidance. Additionally, we incorporate a boolean switch to control the obstacle avoidance behavior of the pedipulating foot. The following section details the training process, simulation environment, and our Markov Decision Process (MDP) design for RL training.

\subsection{Training Overview}
We train our RL policy in three stages~(\figref{fig:stages}):
\begin{enumerate}
    \item Initially, the agent learns obstacle-free pedipulation with a performance-based curriculum on the command space (\figref{fig:stages}-A and \figref{fig:stages}-B), following our previous work \cite{arm2024pedipulate}.
    \item Once the command space curriculum has converged, we introduce obstacles (\secref{sec:obstacle_scenarios}) and collision-avoidance rewards (\secref{sec:rewards}).
    \item Finally, to overcome the sim-to-real perception gap, we add noise to the perceptive observations, emulating the noise introduced by the perception pipeline used during deployment on the ANYmal quadruped (\secref{sec:sim2real}).
\end{enumerate}

We separate Stages 1 and 2 to speed up training, as adding obstacle avoidance from the start makes the already hard problem of obstacle-free pedipulation more difficult and slows down training progress. 

Explicitly modeling the different types of obstacles encountered in real life is infeasible due to their number and varying geometries. We hypothesize that with a small set of obstacles arranged in a well-constructed set of obstacle scenarios, the agent can learn a generalizable obstacle-avoiding behavior. To this end, we train the policy in parallel on all five scenarios by randomly selecting one scenario for each of the 4096 environments.

We add Stage 3 to improve the sim-to-real transfer, as the elevation map on the robot is subject to noise, outliers, and occlusions.
Training of a policy on all three stages takes roughly 6 days on an Nvidia RTX 4090.

\subsection{Simulation Environment}
We design the simulation environment in NVIDIA Isaac Lab~\cite{mittal2023orbit}. The simulation scene comprises rigid bodies for obstacles and the articulated rigid robot body.

\subsubsection{Obstacles}

We use cuboids as obstacles, with size and mass randomization according to \tabref{tab:obstacle_rand}.

Previous works targeting obstacle avoidance integrate obstacles into the static terrain \cite{miki2024confined, Frey_2022}, which is computationally efficient as it avoids the computation of the obstacle dynamics.
In the context of pedipulation, however, where moving obstacles should be possible when desired (\secref{sec:commands}), training on static obstacles only would prevent the policy from performing interactive behaviors such as pushing or reorientation of obstacles during deployment. Thus, we use movable obstacles, which increase training times but provide representative experiences of collision dynamics.

\begin{table}[t]
    \centering
    \normalfont
    \caption{Size and mass randomization of the three obstacles used during training. $\mathcal{U}$ denotes a uniform distribution.}
    \begin{tabular}{lcc}
        \toprule
        \textbf{Property} & \textbf{Obstacle 1} & \textbf{Obstacles 2,3} \\
        \midrule
        \rowcolor[HTML]{EFEFEF} Width & $\mathcal{U}(0.6m, 1.0m)$  & $0.6m$ \\
        Length & $0.5m$ & $0.6m$ \\
        \rowcolor[HTML]{EFEFEF} Height & $\mathcal{U}(0.2m, 1.0m)$ & $\mathcal{U}(0.5m, 1.2m)$\\ 
        Mass & $\mathcal{U}(10kg, 30kg)$ & $\mathcal{U}(10kg, 30kg)$ \\
        \bottomrule
    \label{tab:obstacle_rand}
    \end{tabular}
    \vspace{-30pt}
\end{table}

\subsubsection{Obstacle Scenarios}
\label{sec:obstacle_scenarios}

To encourage the policy to learn generalizable obstacle avoidance, we construct five obstacle scenarios (\figref{fig:stages}, Stage 2). We construct two scenarios using a single obstacle (\tabref{tab:obstacle_rand}, Obstacle 1), focusing on close and far-range obstacle avoidance (\figref{fig:stages}-C and \figref{fig:stages}-D, respectively). 
Our third scenario (\figref{fig:stages}-E) has two obstacles (2 and 3) with a gap wide enough for the robot to spawn in between, which encourages the policy to act conservatively when constrained on both sides. Similarly, in the next environment (\figref{fig:stages}-F), we use two obstacles (2 and 3) to create a tight gap, into which the robot can only reach with its foot. Finally, using all three obstacles, we construct a scenario that features two gaps to reach through and position the obstacles such that there is enough space for the robot to turn on the spot to reach commands behind it (\figref{fig:stages}-G).

\subsubsection{Robot}
\label{sec:robot}
In the context of obstacle avoidance, it is important to choose an appropriate robot collision model that balances accuracy and computational efficiency (\figref{fig:collision_model}).
The feet, which frequently encounter obstacles, are modeled accurately, while the base, which should rarely see collisions, is approximated with a cuboid.
Conservative collision bodies for the knees and hips help the policy avoid self-collisions and challenging poses, such as reaching between its legs instead of rotating the base to track foot position commands behind the robot.
Finally, we choose the right front foot (RF\_foot) to perform pedipulation in this work.

\subsubsection{Initial Spawning Location}

We randomize the robot's spawn position near obstacles to increase the likelihood of the policy having to avoid obstacles given random commands. We find a flat patch large enough to spawn within, to ensure the robot does not spawn in collision with obstacles. To this end, we sample a random position and orientation from the spawn space and verify that the outline of a 1.1m by 0.8m rectangle at that position and orientation is sufficiently flat. If a sampled position is invalid, we resample until the position is valid. This approach allows for scenarios where the robot spawns between obstacles (e.g.~\figref{fig:stages}-E).

\subsection{Observations}
\label{sec:observations_and_actions}

\begin{table}[h]
    \centering
    \vspace{5pt}
    \caption{Observation Terms Summary. Terms expressed in the robot base frame carry the subscript $\mathcal{B}$. The height scan is a stacked vector of height values $h$ sampled from an $l$ by $k$ grid and clipped to the range $[0.0, 1.0]$.}
    \normalfont
    \begin{tabular}{l@{\hspace{-12pt}}cc}
        \toprule
        \textbf{Observation Term Name} & \textbf{Definition} & \textbf{Noise} \\
        \midrule
        \rowcolor[HTML]{EFEFEF}Base Linear Velocity & ${}_{\mathcal{B}}\mathbf{v}$  & $\mathcal{U}(-0.1, 0.1)$ \\
        Robot Base Angular Velocity & ${}_{\mathcal{B}}\mathbf{\Omega}$ & $\mathcal{U}(-0.2, 0.2)$ \\
        \rowcolor[HTML]{EFEFEF}Base Projected Gravity & ${}_{\mathcal{B}}\mathbf{g}$ & $\mathcal{U}(-0.05, 0.05)$ \\
        Robot Joint Positions & $\mathbf{q}_j$ &  $\mathcal{U}(-0.01, 0.01)$ \\
        \rowcolor[HTML]{EFEFEF}Joint Velocities & $\dot{\mathbf{q}_j}$ &  $\mathcal{U}(-1.5, 1.5)$ \\
        Foot Position Command & $\mathbf{p}_{foot}^{*}$ & - \\
        \rowcolor[HTML]{EFEFEF} Previous Actions & $\mathbf{q}_{j}^{*}$ & - \\
        Height Scan & $\begin{bmatrix} h_{1,1}, h_{1,2}, ..., h_{l, k} \end{bmatrix}$ & See \figref{fig:stages}-I \\
        \rowcolor[HTML]{EFEFEF}Contact Switch & $ b_{switch} \in \{0, 1\}$ & - \\
        \bottomrule
    \label{tab:obs}
    \end{tabular}
    \vspace{-5pt}
\end{table}

\begin{figure}[h!]
    \centering
    \begin{minipage}[t]{0.46\linewidth}
        \includegraphics[width=\linewidth, trim={6cm, 1cm, 4cm, 1cm}, clip]{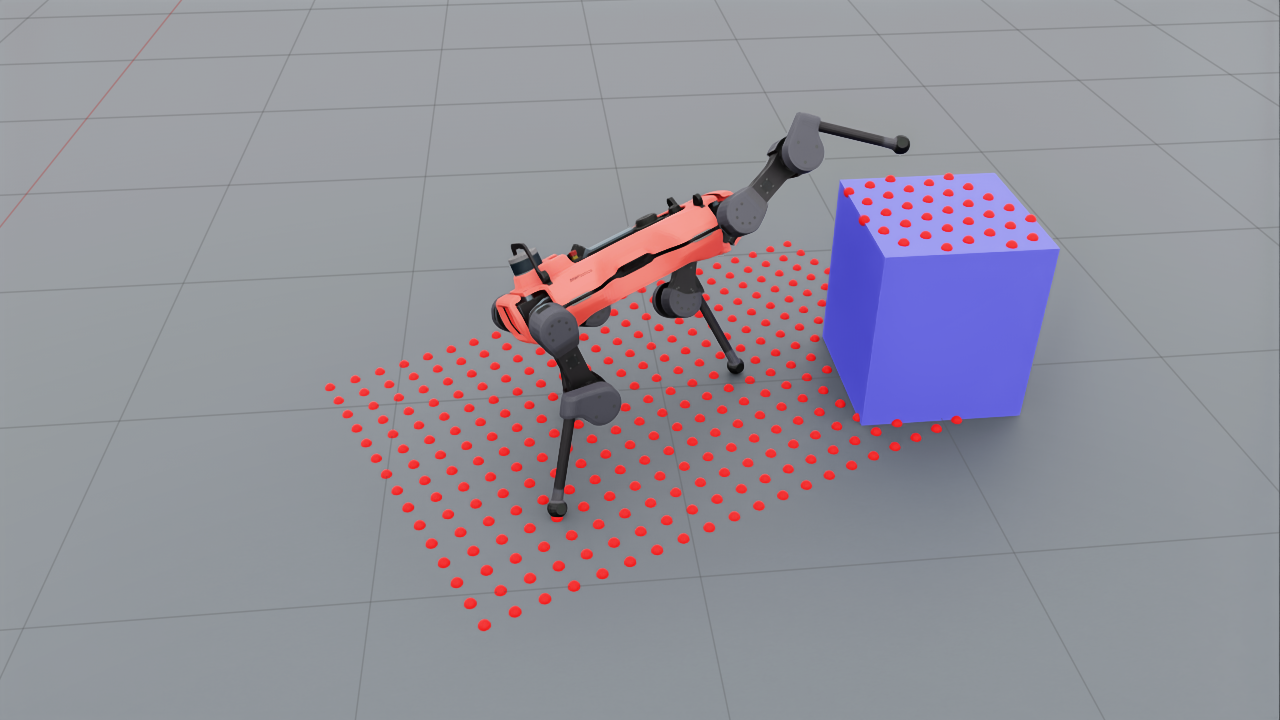}
        \caption{We use a height scan of size $\SI{2.4}{\meter} \times \SI{1.6}{\meter}$ with a resolution of $\SI{0.1}{\meter}$. The grid is shifted to the front by $\SI{0.2}{\meter}$ to cover the reach of the pedipulating foot.}
        \label{fig:perception}
    \end{minipage}
    \hspace{4pt}
    \begin{minipage}[t]{0.46\linewidth}
        \includegraphics[width=\linewidth, trim={5cm, 1cm, 7cm, 2cm}, clip]{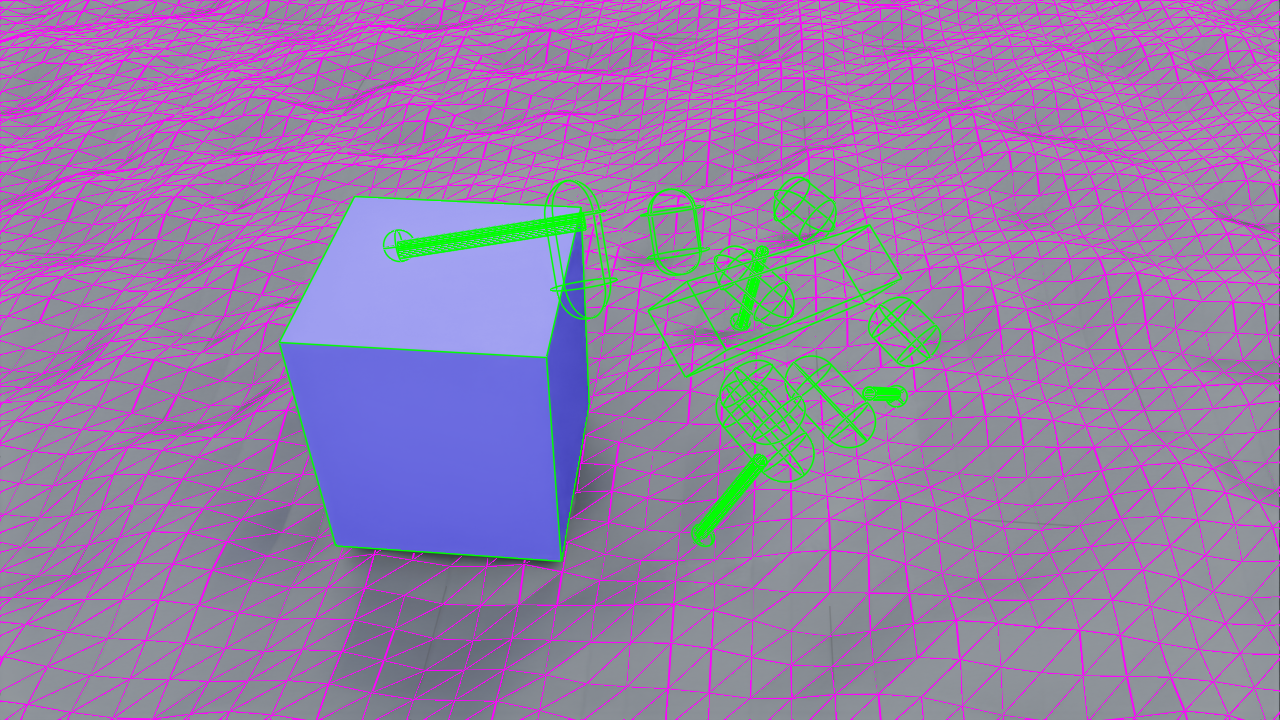}
        \caption{We use a detailed collision model for the robot and train on rough ground to encourage the policy to take higher steps. The obstacles are simple cuboids.}
        \label{fig:collision_model}
    \end{minipage}
    \vspace{-5pt}
\end{figure}
We choose proprioceptive observations that are available on the hardware through the state estimator, similar to the prior work~\cite{arm2024pedipulate}. The observations include the actions $\mathbf{q}^{*}$, which we interpret as joint position offsets relative to the default joint positions.

In addition, we provide perceptive observations in the form of a 2.5D height scan (\figref{fig:perception}) previously used in locomotion tasks \cite{miki2022learning,miki2022elevation}, which keeps the dimension of the observation vector low. A low-dimensional observation vector reduces training times and complexity for the relatively complex task of learning pedipulation and obstacle avoidance end-to-end.

\subsection{Commands}
\label{sec:commands}
Commands define the goal of the task and need to be provided by the human operator or a high-level planner during deployment. Following previous work on pedipulation~\cite{arm2024pedipulate}, we only command the foot position and allow the rest of the whole-body control to emerge during training. During training, the foot position commands are sampled uniformly from the current command space. In stage one, we apply a command space curriculum. In the subsequent stages, we provide separate command spaces for each obstacle scenario (\figref{fig:stages}). Over a single episode, we resample the commands three times with 6 seconds between each resample.

While whole-body obstacle avoidance is desirable, we additionally need the option to allow the pedipulating foot to make contact with the environment to solve manipulation tasks, such as pushing doors open or moving objects.
To this end, we provide an additional boolean command $b_{switch}$. During training, this switch is set randomly for every episode and determines whether penalties on the pedipulating foot's contact forces are applied (\tabref{tab:rewards}). The policy then learns to avoid obstacles when the switch is off and to allow contacts when it is on. Note that the remaining bodies of the robot always have contacts penalized and should thus always avoid obstacles.

\begin{table}[t]
    \centering
    \vspace{5pt}
    \caption{Reward Terms Summary. $\mathbf{F}$ denotes the sum of external forces on a body. Superscript $1$: only applied in Stages 2 and 3. Superscript $2$: only applied if the contact switch $b_{switch}$ is zero. Subscript $\mathcal{B}$: value expressed in the robot base frame.}
    \begin{tabular}{l@{\hspace{-15pt}}c@{\hspace{5pt}}c}
        \toprule
        \textbf{Reward Term Name} & \textbf{Definition} & \textbf{Weight} \\
        \midrule

        \rowcolor[HTML]{EFEFEF} Command Tracking & $exp\Big(-\frac{\left\lVert \mathbf{p}_{foot}^{*} - \mathbf{p}_{foot}\right\rVert}{0.8}\Big)$ & \num{14.0}\\ 
        Termination & 
        if $\left\lVert \mathbf{F}_{base}\right\rVert > 1.0$ & \num{-500.0} \\
        \rowcolor[HTML]{EFEFEF} Base Linear Velocity $z$ & ${}_{\mathcal{B}}\dot{z}^2 $ & \num{-2.0}\\
        Base Angular Velocity $xy$ & $\left\lVert {}_{\mathcal{B}}\boldsymbol{\omega}_{xy}^2\right\rVert$ & \num{-0.05} \\
        \rowcolor[HTML]{EFEFEF}Torques & $\boldsymbol{\tau}^{T}\boldsymbol{\tau}$ & \num{-2.0e-5}\\
        Joint Velocities & $\dot{\mathbf{q}}^{T}\dot{\mathbf{q}}$ & \num{-0.04} \\
        \rowcolor[HTML]{EFEFEF} Joint Accelerations & $\ddot{\mathbf{q}}^{T}\ddot{\mathbf{q}}$ & \num{-2.5e-7} \\
        Action Rate & $\dot{\mathbf{q}}^{*T}\dot{\mathbf{q}}^{*}$ & \num{-0.02}\\

        \rowcolor[HTML]{EFEFEF} Contact Events \hspace{-20pt} & $\sum\limits_{i \in \{robot\_bodies\}} (\left\lVert \mathbf{F}_{i}\right\rVert > 0.1)$ & \num{-2.0}\\        
        \midrule
        \midrule
        \textbf{Contact Events for Obstacle Avoidance} \hspace{-40pt} & \hspace{-40pt} & \\
        \midrule
        
        \rowcolor[HTML]{EFEFEF} Pedipulating Foot${}^{1, 2}$  &  $\left\lVert \mathbf{F}_{RF\_foot}\right\rVert > 0.001 $
        & \num{-80.0}\\
        
        Remaining Feet${}^{1}$ & \hspace{-20pt}$\sum\limits_{i  \in \{remaining\_feet\}}(\left\lVert \mathbf{F}_{i}\right\rVert > 0.001)$ & \num{-20.0}\\
        
        \rowcolor[HTML]{EFEFEF} Hips and Knees${}^{1} \hspace{-20pt} $ & $\sum\limits_{i \in \{knees\} \cup \{hips\}} \left\lVert \mathbf{F}_{i}\right\rVert > 0.001$ & \num{-40.0}\\

        \midrule
        \textbf{Contact Forces for Obstacle Avoidance} \hspace{-40pt} &\hspace{-40pt} & \\
        \midrule
        
        \rowcolor[HTML]{EFEFEF} Pedipulating Foot${}^{1,2}$& 
        $ \left\lVert \mathbf{F}_{RF\_foot} \right\rVert$
          & \num{-40.0}\\

        Remaining Feet${}^{1}$& 
        $\sum\limits_{i \in \{remaining\_feet\}}  \left\lVert \mathbf{F}_{i}\right\rVert$
        & \num{-0.2}\\

        \rowcolor[HTML]{EFEFEF} Hips and Knees${}^{1}$& 
        $\sum\limits_{i \in \{hips\} \cup \{knees\}}  \left\lVert \mathbf{F}_{i}\right\rVert$
        & \num{-0.2}\\
        \bottomrule
    \end{tabular}
    \renewcommand{\arraystretch}{2.0}
    \label{tab:rewards}
    \vspace{-6pt}
\end{table}

\subsection{Rewards}
\label{sec:rewards}
In the first training stage (\figref{fig:stages}), we use a dense foot tracking reward, penalties for undesired contacts and excessive base movement, and regularization rewards following~\cite{arm2024pedipulate}. Specifically, we regularize the joint positions $\mathbf{q}$, their first and second derivatives, torques $\boldsymbol{\tau}$, and the rate of change of the actions $\mathbf{\dot{q}^{*}}$.
We terminate the episode when the external force on the base exceeds a threshold, which indicates that the robot fell over. Smaller forces on the base, which can occur due to self-collisions, are only penalized.
When introducing the obstacles in the second training stage, we add a contact reward structure with individually tuned weights for different parts of the robot's body (\tabref{tab:rewards}, Contact Events/Forces for Obstacle Avoidance).
To provide a smoother transition of the reward structure, we apply a curriculum on these rewards: They linearly increase from zero to their final weight over 5000 steps.
We penalize both contact force magnitudes and contact events to prevent two exploitation strategies:
\begin{enumerate}
    \item Without penalizing contact events, the policy could learn to push an obstacle with minimal force, which would be a limited penalty in return for an exponential command tracking reward.
    \item Without penalizing the force magnitudes, on the other hand, the policy could push the obstacle out of the way with a short but high-intensity push. 
\end{enumerate}
We apply this idea to all collision bodies of the robot to avoid exploitation strategies like pushing obstacles away with the hips, for example, which receive lower penalties than the pedipulating foot.
Note that we choose a minimum force threshold for the contact events to avoid false positives due to numerical inaccuracies. 
We only penalize the non-pedipulating feet for contacts with obstacles but not the ground, as those forces should clearly be allowed.
Finally, we give the contact rewards for the pedipulating foot based on the contact switch command $b_{switch}$ (\secref{sec:commands}). All listed rewards were tuned manually by inspecting the performance of a trained policy and adjusting weights accordingly before retraining.

\subsection{Sim-to-Real Transfer}
\label{sec:sim2real}
On hardware, we sample the height scan from an elevation map created by the proprietary ANYmal software based on the depth cameras on the four sides of the robot (\figref{fig:stages}-I). We tuned the elevation mapping parameters to be conservative about occlusions and not assume thin-walled objects where the top of the object cannot be seen (e.g.~\figref{fig:real_bin}).

This map, however, is subject to noise and artifacts, which are not present in the idealized simulation. 
To enable the sim-to-real transfer, we increase the randomization of the height scan observations in the third training stage. With likelihood 0.05, we set points to random values sampled from $\mathcal{U}(0.0m, 1.3m)$, emulating the artifacts and edge noise in the elevation map. With likelihood 0.3, we set points to zero, emulating the effects of occlusions and blind spots. We also add drift and random noise.

In addition, over all three stages, we add the following randomizations.
Following our previous work \cite{arm2024pedipulate}, we choose the ground to be a height field (see \figref{fig:collision_model}) uniformly sampled from $\mathcal{U}(-0.05m, 0.05m)$, which encourages the policy to take higher steps.
To make the policy robust to unseen disturbances, we apply external forces to the robot's base and the pedipulating foot. We also randomize friction parameters and the robot's mass.

\section{Results and Analysis}

We first evaluate obstacle-avoiding performance in simulation for ease of repeatability and visualization. Secondly, we demonstrate that these skills transfer to hardware by performing experiments on the ANYmal quadruped \cite{ANYmal-D}.

\subsection{Simulation Testing}

\begin{figure}[t]
    \centering
    \vspace{6pt}
    \begin{overpic}[width=\linewidth, trim={0cm 0cm 0cm 0cm}, clip]{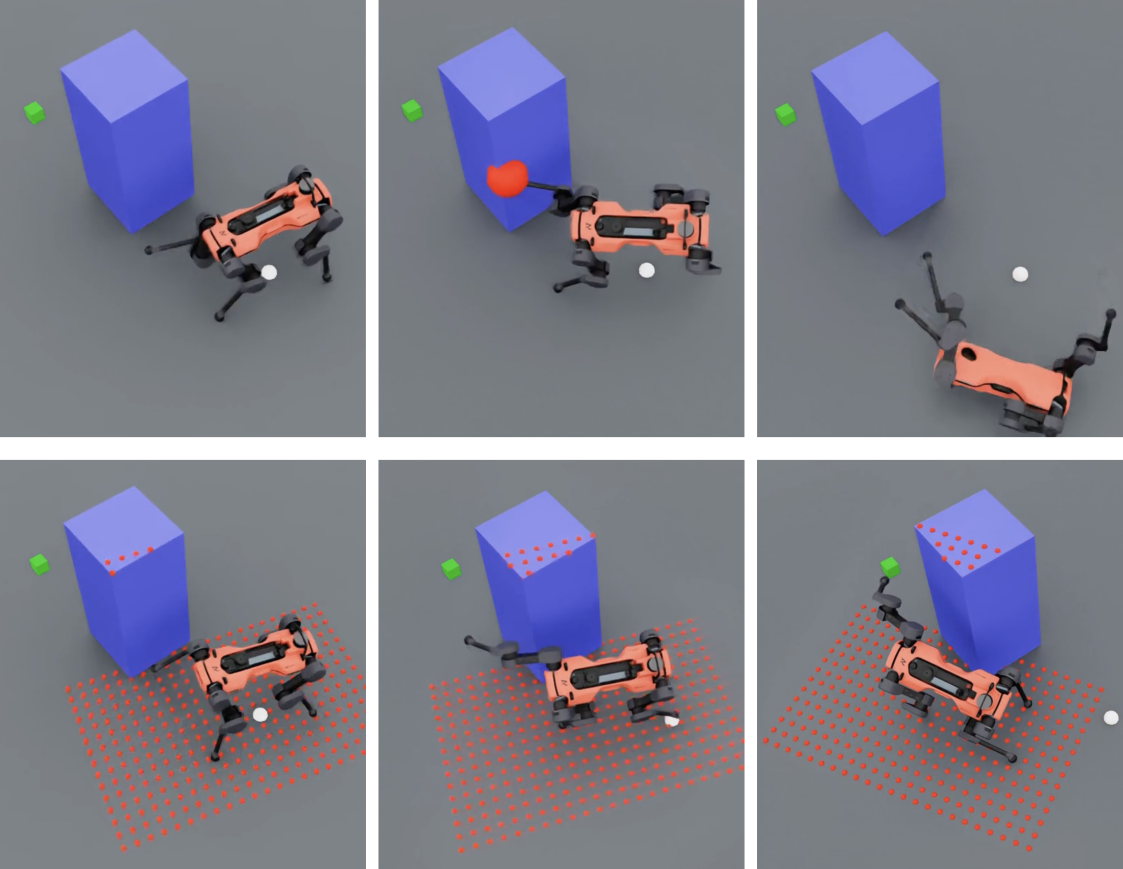}
        \put(3,42){\onimagetext{A}}
        \put(3,3){\onimagetext{B}}
    \end{overpic}
    \caption{
    The blind pedipulation policy \cite{arm2024pedipulate} tries to reach the foot position command (green). The obstacle in the path of the foot leads to a collision (red) and eventually to a critical failure of the policy (A). Our perceptive pedipulation policy can reach around a corner while avoiding collisions (B).}
    \label{fig:perceptive_around_corner}
    \vspace{-6pt}
\end{figure}

\begin{figure*}[t]
    \centering
    \vspace{6pt}
    \begin{overpic}[width=\textwidth]{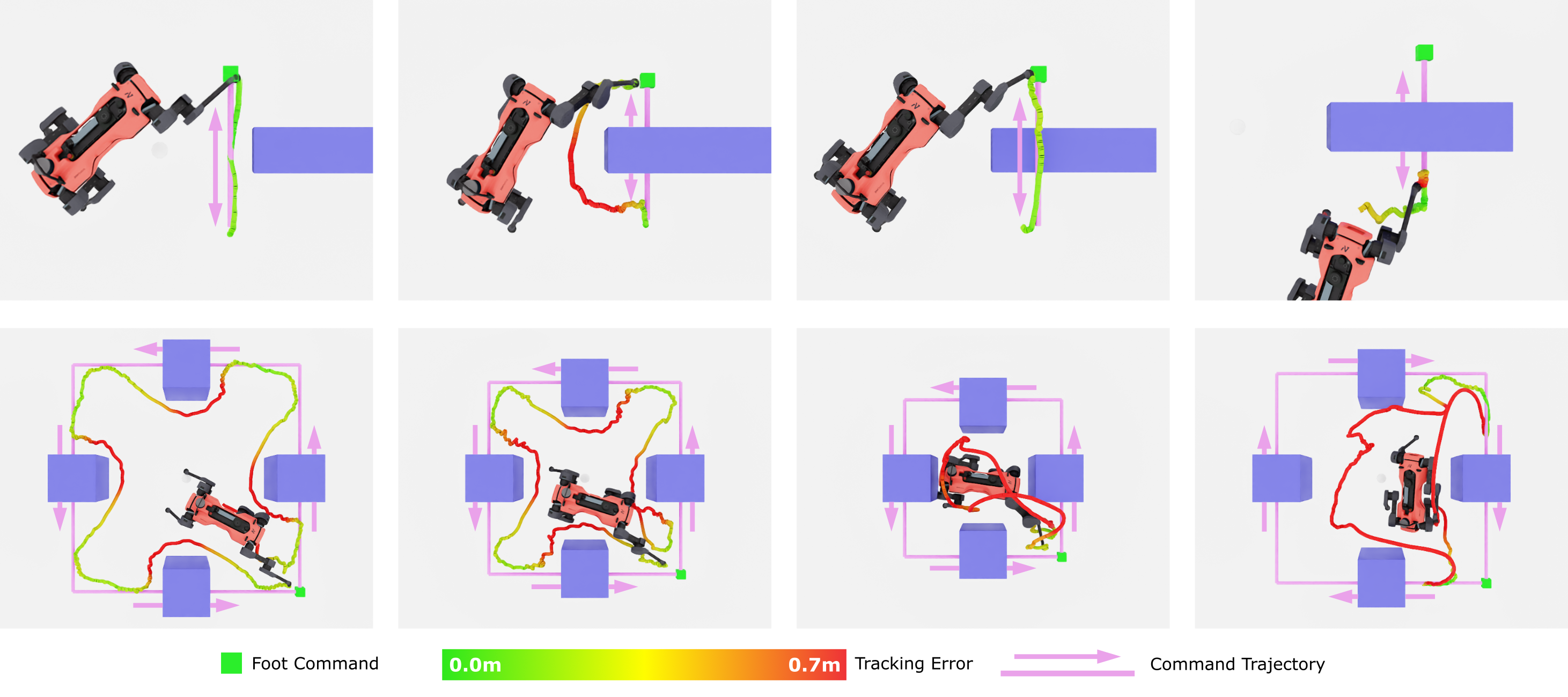}
        \put(20.5,27){\onimagetext{A}}
        \put(46,27){\onimagetext{B}}
        \put(71,27){\onimagetext{C}}
        \put(96.5,27){\onimagetext{D}}
        \put(20.5,6){\onimagetext{E}}
        \put(46,6){\onimagetext{F}}
        \put(71,6){\onimagetext{G}}
        \put(96.5,6){\onimagetext{H}}
    \end{overpic}
    \caption{We evaluate obstacle avoidance in simulation, providing dense foot position command trajectories for single and multiple obstacle scenarios. In the top row, we use a single obstacle and provide commands to move the foot from one side to another. In the bottom row, the robot needs to follow a square ring pattern around four obstacles. While effectively avoiding obstacles in many scenarios, the policy sometimes performs adversely. For instance, getting stuck due to a limited perceptive field of view (D) or insufficient free space to rotate the base freely (G).}
    \label{fig:sim_tests}
\end{figure*}
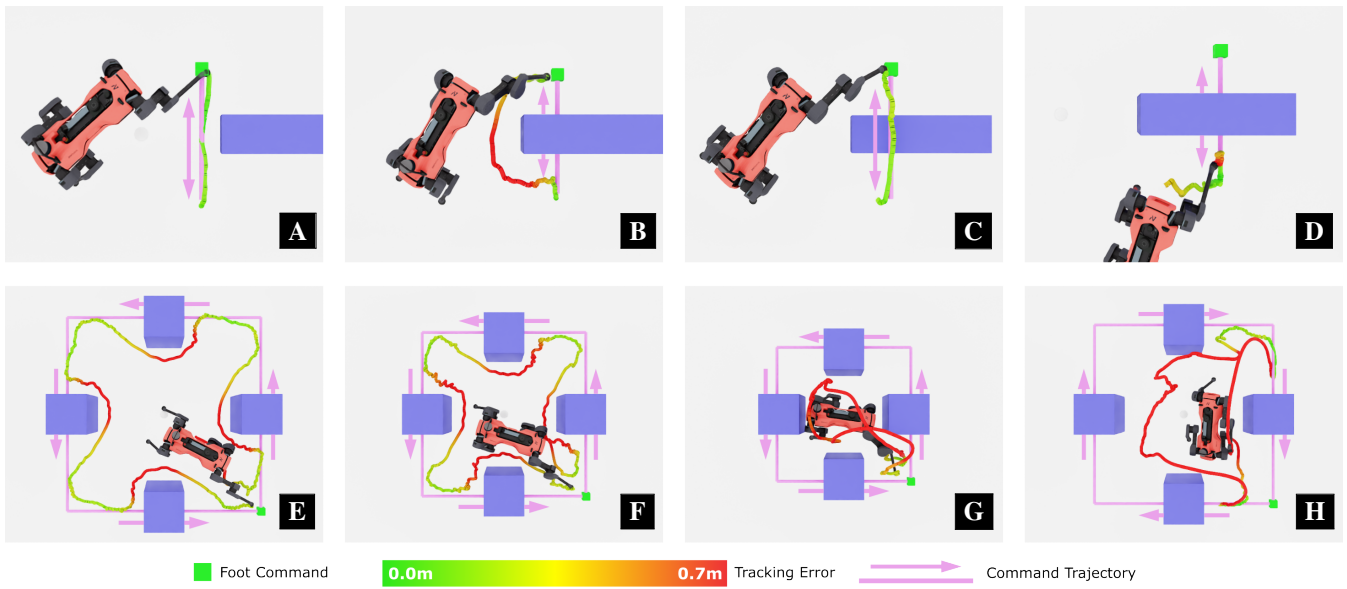

\begin{figure*}[t]
    \centering
    \begin{minipage}{\textwidth}
        \centering
        \begin{overpic}[width=\linewidth]{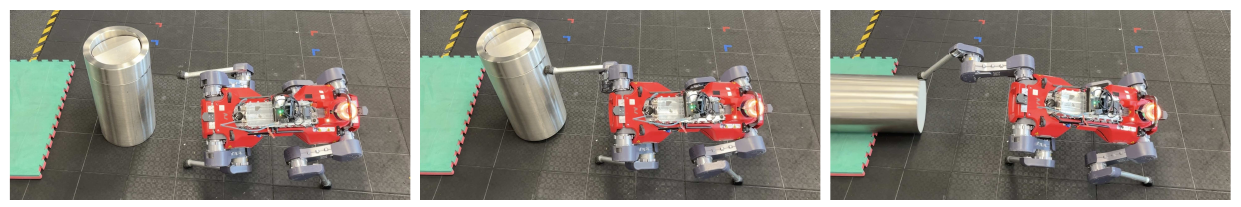}
        \put(1.5,13.0){\onimagetext{A}} %
        \put(34.5,13.0){\onimagetext{B}} %
        \put(67.5,13.0){\onimagetext{C}} %

    \end{overpic}
    \end{minipage}
    \vspace{-20pt}
    \begin{minipage}{\textwidth}
        \centering
        \includegraphics[width=\textwidth, trim={0 0 0 0}, clip]{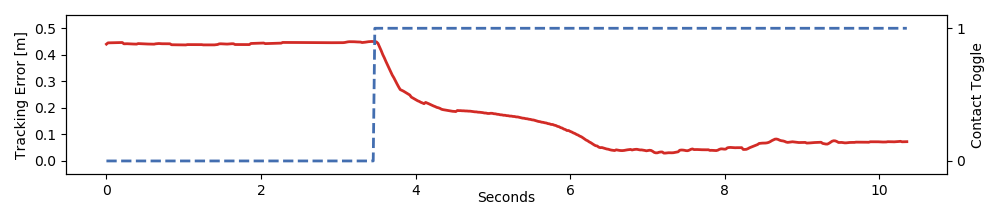 }
    \end{minipage}
    \vspace{5pt}
    \caption{With a constant foot position command inside the obstacle and the contact switch at 0 (A), the robot avoids the obstacle with a large tracking error (red). Once we set the contact switch to 1, the foot moves to the commanded position (B), tipping over the obstacle in the process (C). This switching helps enable interactions with the environment when desired. The final tracking error achieved is \SI{0.073}{\meter}.}
    \label{fig:real_switch}
    \vspace{-10pt}
\end{figure*}

\begin{figure*}[t]
    \centering
    \vspace{5pt}
    \begin{overpic}[width=\linewidth]{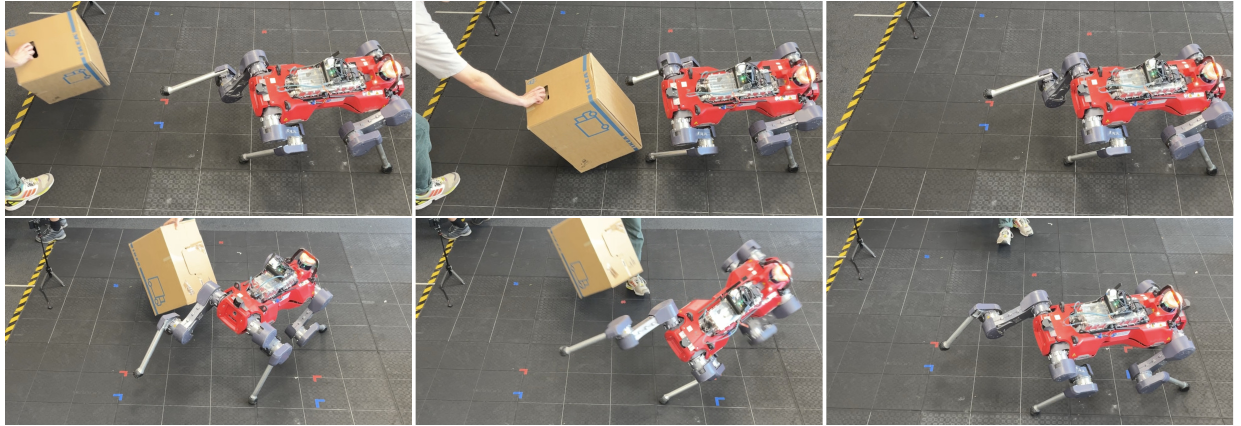}
        \put(1.5,18.5){\onimagetext{A}} %
        \put(1.5,1.5){\onimagetext{B}} %
    \end{overpic}
    \caption{The learned policy is capable of avoiding dynamic obstacles, although it is not explicitly trained for them in simulation. With the obstacle approaching from the front, the robot retracts its pedipulating foot to avoid collisions (A). When the obstacle comes from the side, the robot repositions the base and resumes tracking the foot position command (B).}
    \label{fig:dynamic_tests}
    \vspace{-10pt}
\end{figure*}

\begin{figure}[t]
    \centering
    \begin{overpic}[width=\linewidth]{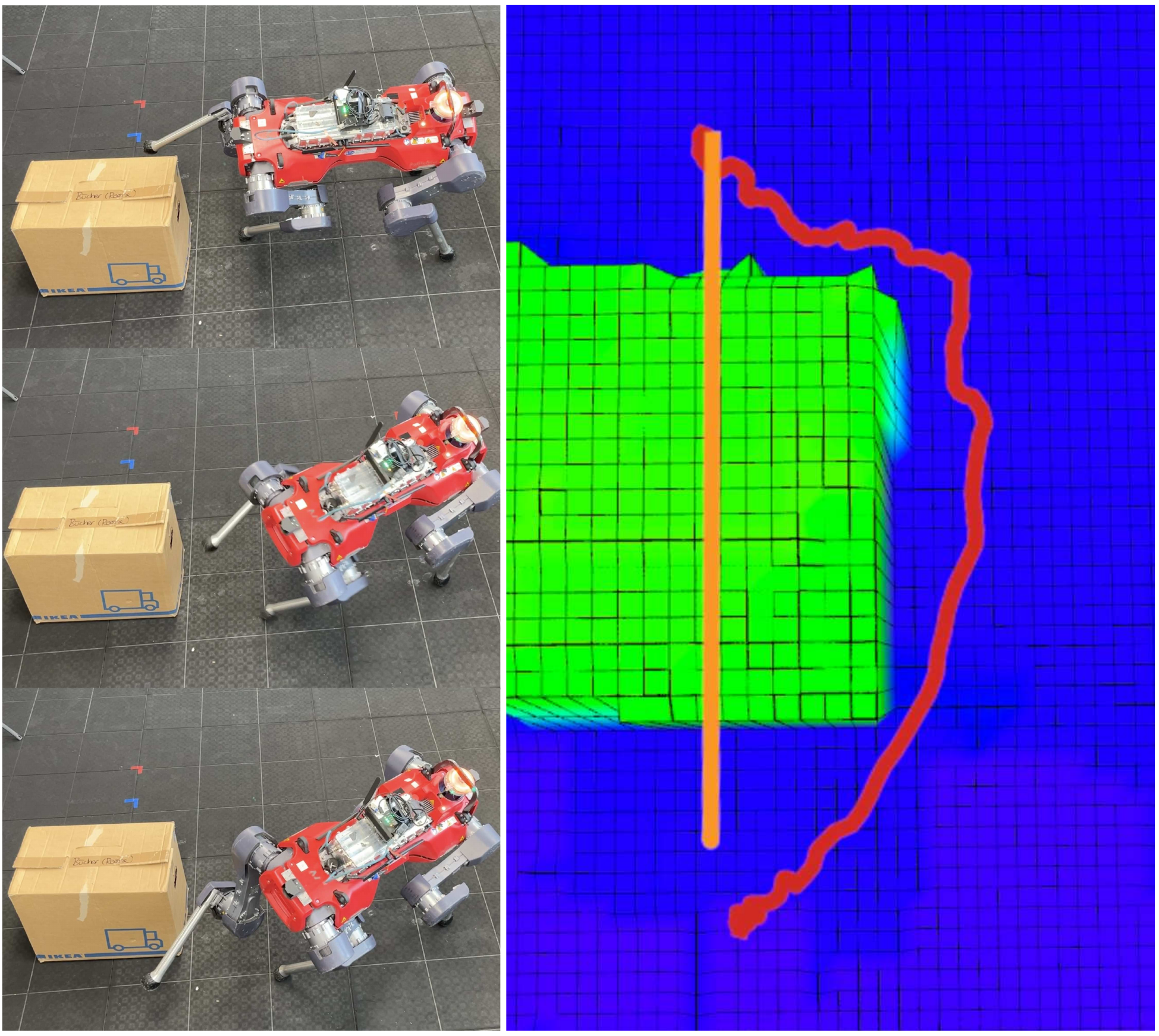}
        \put(2.5,82.0){\onimagetext{A}} %
        \put(2.5,53.0){\onimagetext{B}} %
        \put(2.5,24.0){\onimagetext{C}} %
        \put(70,75.0){\onimagetext{A}} %
        \put(90,45.0){\onimagetext{B}} %
        \put(75,10.0){\onimagetext{C}} %

    \end{overpic}
    \caption{The foot position command (orange) is moved back and forth along a straight line through the cardboard box via teleoperation. The actual foot trajectory (red) goes around the box.}
    \label{fig:real_box}
    \vspace{-8pt}
\end{figure}

\begin{figure}[t]
    \centering
    \begin{overpic}[width=\linewidth]{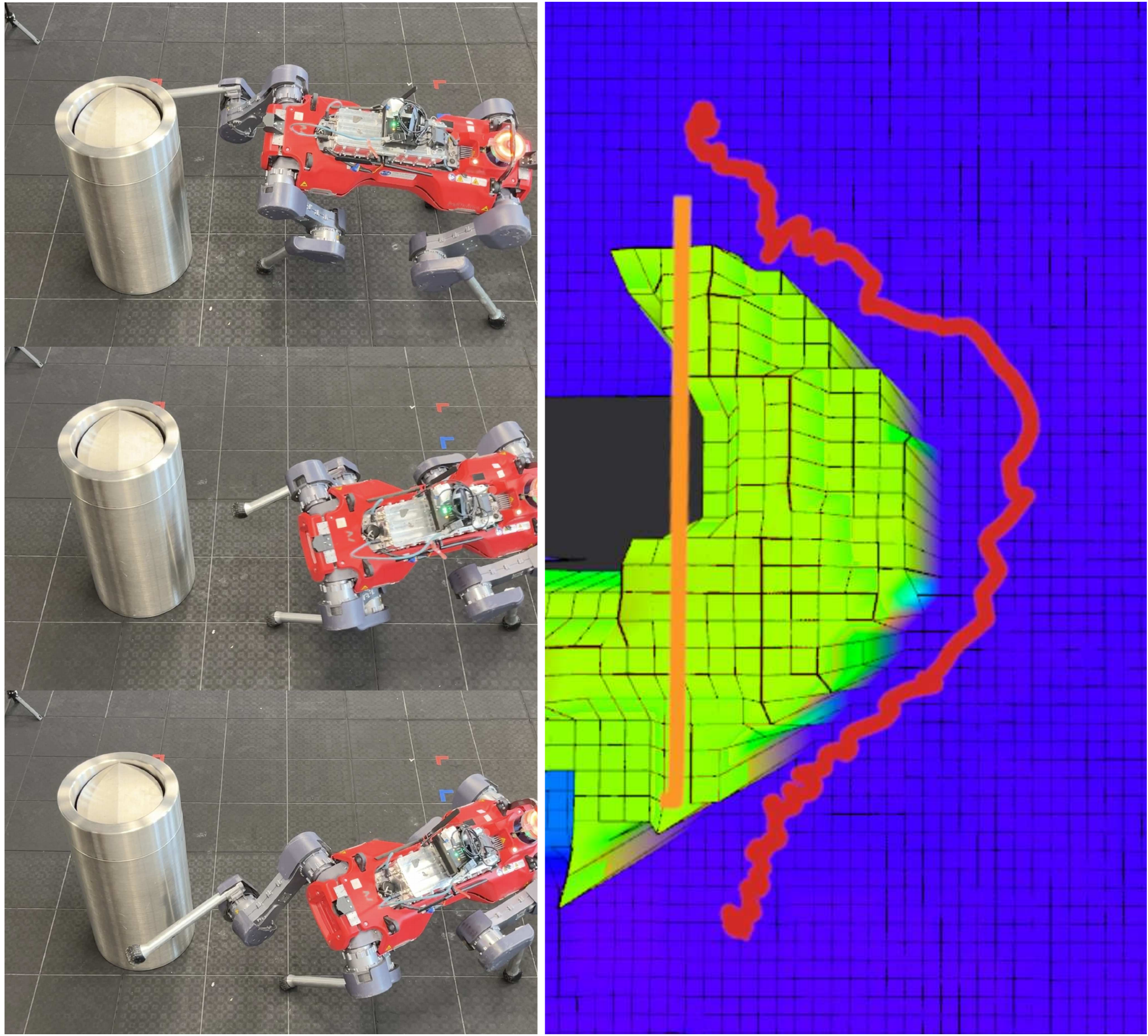}
        \put(2.5,84.0){\onimagetext{A}} %
        \put(2.5,54.0){\onimagetext{B}} %
        \put(2.5,24.0){\onimagetext{C}} %
        \put(70,80.0){\onimagetext{A}} %
        \put(92,50.0){\onimagetext{B}} %
        \put(70,10.0){\onimagetext{C}} %
    \end{overpic}
    
    \caption{The foot position command (orange) is moved back and forth along a straight line through the metal bin via teleoperation.
    The actual foot trajectory (red) goes around the bin, which has a geometry not seen during training.
    We handle holes in the elevation map conservatively by replacing them with large height values.
    }
    \label{fig:real_bin}
    \vspace{-5pt}
\end{figure}

\subsubsection{Reaching Around a Corner}

We demonstrate the benefits of our perceptive pedipulation policy by comparing it to our previous blind policy \cite{arm2024pedipulate} in the simple scenario of reaching around a corner (see \figref{fig:perceptive_around_corner}). Due to the dense tracking reward (see \secref{sec:rewards}), the blind policy (A) moves the pedipulating foot to the foot position command on a direct path.  As an obstacle is obstructing this path, the foot collides with the corner of the obstacle, which leads to the robot falling over. In contrast, our perceptive policy (B) is aware of the obstacle and navigates around it, taking a longer path to the foot position command, while successfully avoiding collisions.

\subsubsection{Free Space Tracking}

We investigate the effect of the contact switch on the free space tracking accuracy, i.e., when there are no obstacles in the robot's vicinity. We randomly sample 1024 foot position commands from a \SI{2}{\meter} x \SI{2}{\meter} x \SI{1.3}{\meter} command space and average the tracking errors. We get an average tracking error of 0.057m with the switch set to zero. With the switch set to one, we get 0.047m. This experiment indicates that disabling the contact switch does degrade the tracking accuracy, albeit only slightly.

\subsubsection{Single Obstacle}

To test local obstacle avoidance, we first consider a single obstacle (\figref{fig:sim_tests}-A to \figref{fig:sim_tests}-D). We move the foot position command from one side of the obstacle to the other at a constant height of \SI{0.4}{\meter}. As opposed to training, here, we provide a dense trajectory, as this is representative of a teleoperation scenario. 
When the entire command trajectory is out of collision, the robot's foot tracks it as expected.
When the command moves through an obstacle, the foot takes a longer path around the obstacle, which results in a larger tracking error but avoids collisions (\figref{fig:sim_tests}-B). 

When the obstacle's height is lowered enough, the foot will take a path above the obstacle instead of around it, if this path results in a lower tracking error (\figref{fig:sim_tests}-C). This result demonstrates that the policy does indeed consider the height of the obstacle and does not just avoid any regions in the xy-plane that have nonzero height.
Finally, if the trajectory is moved too far into an obstacle, the policy will keep the foot on one side, behaving conservatively while continuing to avoid collisions (\figref{fig:sim_tests}-D). This behavior likely occurs due to the dense reward structure. Going around the obstacle would reduce the reward until the robot can move toward the target again.

Our policy successfully avoided self-collisions and collisions with the environment in all four tests. In each test, the command moved back and forth twice. This example also shows that training on relatively wide obstacles generalizes to thinner obstacles.

\subsubsection{Multiple Obstacles}

The final simulation experiment tests the collision-avoidance behavior near multiple obstacles (\figref{fig:sim_tests}-E to \figref{fig:sim_tests}-H).
When the obstacles are far apart (\figref{fig:sim_tests}-E), the policy uses its pedipulation capabilities and tripod gait to track the command trajectory, avoiding obstacles when it encounters them.
Moving the obstacles closer together (\figref{fig:sim_tests}-F), the policy first displays a similar behavior as before until we reach a limit at a gap size of \SI{1.3}{\meter}, where rotating the base would likely lead to collisions (\figref{fig:sim_tests}-G). At this point, the policy stops turning the base and tracks the foot position commands behind it by reaching behind its back. In this situation, some low-force collisions occur with the back side of the robot base.

Changing the trajectory of the foot position command from counterclockwise to clockwise significantly degrades tracking performance (\figref{fig:sim_tests}-H). Here, the robot tends to reach behind its back even when there is enough space to turn the base. From this pose, the policy struggles to rotate the base towards the target and return to the default tracking behavior. One possible reason for this behavior is that the policy only tracks points but not complete trajectories. Moving the foot to the back is the easiest way to reach the point behind the robot, as seen in the bottom right of~\figref{fig:sim_tests}-H. However, this position is unfavorable for tracking the rest of the trajectory. Observing trajectories instead of individual points could mitigate this issue.

\subsection{Hardware Experiments}

\subsubsection{Contact Switch}

First, we demonstrate the functionality of the contact switch with a simple example (\figref{fig:real_switch}).
When we set the switch to obstacle avoidance and move the foot position command inside the obstacle, the policy successfully avoids it. 
Once we toggle the switch to allow contacts with the pedipulating foot, the robot pushes the obstacle out of the way to reach the commanded position.

\subsubsection{Single Obstacle}
The obstacle-avoiding behaviors demonstrated in simulation transfer effectively to real hardware. The policy can track foot positions through obstacles like boxes (\figref{fig:real_box}) while avoiding collisions. The robot repositions its base using a tripod gait when the foot position command is out of reach or obstructed. We successfully repeated this experiment five times.

Notably, we perform the same tests on a round obstacle (\figref{fig:real_bin}), which is not part of the training scenarios. The policy’s success in avoiding this new geometry highlights its generalization capabilities. 
We believe this works because the policy does not just recognize and avoid the rough shapes of obstacles. Instead, the agent learns to avoid the individual columns of space encoded in the height scan if they are occupied. We infer that the policy considers each cell in the grid individually, enabling it to precisely avoid obstacles regardless of their shape. Note that this excludes very small obstacles, which we consider perception artifacts.

\subsubsection{Dynamic Obstacles}

Even though the obstacles do not move actively in simulation, the policy displays a dynamic obstacle-avoiding behavior. It successfully moves the pedipulating leg away from an incoming obstacle and returns to the commanded foot position once it is removed (\figref{fig:dynamic_tests}-A). The elevation map update rate limits the reaction speed of the controller.  The robot avoids obstacles coming in from the side by repositioning its base (\figref{fig:dynamic_tests}-B). This behavior, however, only works for obstacles on the robot's right side.

\section{Conclusion}

This paper addressed the challenge of local collision avoidance for legged robots performing pedipulation. We developed a deep reinforcement learning policy incorporating a 2.5D height scan for perceptive information, enabling obstacle-avoiding whole-body control while tracking foot position commands.

Through simulation and hardware experiments, we demonstrated that our policy generalizes from a minimal set of obstacle scenarios during training to previously unseen obstacle geometries. Even though only trained on non-moving obstacles, the policy displays dynamic obstacle-avoiding behavior. The discretized nature of our perception representation allows the policy to generalize to unseen scenarios. However, it limits the set of observable obstacle geometries.

Human operators or high-level planners could easily and safely use our controller for pedipulation tasks near obstacles, as it tackles both pedipulation and obstacle avoidance end-to-end. Our contact switch additionally provides the option to switch between whole-body obstacle avoidance - including the foot - and using the foot to interact with the environment.
We see this work as a first step towards robust mobile pedipulation in obstacle-rich and unknown environments. 

\section{Limitations and Future Work}
\label{sec:discussion}

While our can perform obstacle-avoiding foot command tracking repeatedly, there are some corner cases that the controller does not handle well. These include obstacles directly under the base or confined spaces that are not seen during training. The choice of perception representation is limited to structures that can be represented in 2.5D and that are larger than the step size of the height scan grid. 

Moving from a 2.5D elevation map to a 3D representation is desirable to overcome these limitations.

\section{Acknowledgements}

We thank Ren\'{e} Zurbr\"{u}gg for assisting with the simulation implementation of the ray-cast sensor. Additionally, we thank Andrei Cramariuc for his feedback.

\bibliography{root}

\begin{thebibliography}{10}

\bibitem{miki2022learning}
T.~Miki, J.~Lee, J.~Hwangbo, L.~Wellhausen, V.~Koltun, and M.~Hutter, ``Learning robust perceptive locomotion for quadrupedal robots in the wild,'' {\em Science Robotics}, vol.~7, no.~62, p.~eabk2822, 2022.

\bibitem{rudin2022learningwalkminutesusing}
N.~Rudin, D.~Hoeller, P.~Reist, and M.~Hutter, ``Learning to walk in minutes using massively parallel deep reinforcement learning,'' in {\em Proceedings of the 5th Conference on Robot Learning}, vol.~164 of {\em Proceedings of Machine Learning Research}, pp.~91--100, 2022.

\bibitem{fu2022deep}
Z.~Fu, X.~Cheng, and D.~Pathak, ``Deep whole-body control: Learning a unified policy for manipulation and locomotion,'' in {\em Conference on Robot Learning ({CoRL})}, 2022.

\bibitem{Mittal_2022}
M.~Mittal, D.~Hoeller, F.~Farshidian, M.~Hutter, and A.~Garg, ``Articulated object interaction in unknown scenes with whole-body mobile manipulation,'' in {\em 2022 IEEE/RSJ International Conference on Intelligent Robots and Systems (IROS)}, IEEE, Oct. 2022.

\bibitem{arm2024pedipulate}
P.~Arm, M.~Mittal, H.~Kolvenbach, and M.~Hutter, ``Pedipulate: Enabling manipulation skills using a quadruped robot's leg,'' in {\em 2024 IEEE International Conference on Robotics and Automation (ICRA)}, 2024.

\bibitem{cheng2023legs}
X.~Cheng, A.~Kumar, and D.~Pathak, ``Legs as manipulator: Pushing quadrupedal agility beyond locomotion,'' in {\em Conference on Robot Learning ({CoRL})}, 2023.

\bibitem{he2024learning}
Z.~He, K.~Lei, Y.~Ze, K.~Sreenath, Z.~Li, and H.~Xu, ``Learning visual quadrupedal loco-manipulation from demonstrations,'' {\em arXiv preprint arXiv:2403.20328}, 2024.

\bibitem{chen_2022}
L.~Chen, Z.~Jiang, L.~Cheng, A.~C. Knoll, and M.~Zhou, ``Deep reinforcement learning based trajectory planning under uncertain constraints,'' {\em Frontiers in Neurorobotics}, vol.~16, 2022.

\bibitem{chiu2022collisionfree}
J.~Chiu, J.-P. Sleiman, M.~Mittal, F.~Farshidian, and M.~Hutter, ``A collision-free mpc for whole-body dynamic locomotion and manipulation,'' {\em 2022 International Conference on Robotics and Automation (ICRA)}, pp.~4686--4693, 2022.

\bibitem{shi2020circus}
F.~Shi, T.~Homberger, J.~Lee, T.~Miki, M.~Zhao, F.~Farshidian, K.~Okada, M.~Inaba, and M.~Hutter, ``Circus anymal: A quadruped learning dexterous manipulation with its limbs,'' in {\em 2021 IEEE International Conference on Robotics and Automation (ICRA)}, p.~2316–2323, 2021.

\bibitem{jeon2023learning}
S.~G. Jeon, M.~Jung, S.~Choi, B.~Kim, and J.~Hwangbo, ``Learning whole-body manipulation for quadrupedal robot,'' {\em IEEE Robotics and Automation Letters}, vol.~9, pp.~699--706, 2023.

\bibitem{lin2024locoman}
C.~Lin, X.~Liu, Y.~Yang, Y.~Niu, W.~Yu, T.~Zhang, J.~Tan, B.~Boots, and D.~Zhao, ``Locoman: Advancing versatile quadrupedal dexterity with lightweight loco-manipulators,'' {\em arXiv preprint arXiv:2403.18197}, 2024.

\bibitem{bjelonic2021mpcwheeled}
M.~Bjelonic, R.~Grandia, O.~Harley, C.~Galliard, S.~Zimmermann, and M.~Hutter, ``Whole-body mpc and online gait sequence generation for wheeled-legged robots,'' in {\em 2021 IEEE/RSJ International Conference on Intelligent Robots and Systems (IROS)}, pp.~8388--8395, 2021.

\bibitem{villarreal2020mpclocomotion}
O.~Villarreal, V.~Barasuol, P.~M. Wensing, D.~G. Caldwell, and C.~Semini, ``Mpc-based controller with terrain insight for dynamic legged locomotion,'' in {\em 2020 IEEE International Conference on Robotics and Automation (ICRA)}, pp.~2436--2442, 2020.

\bibitem{rudin2022advanced}
N.~Rudin, D.~Hoeller, M.~Bjelonic, and M.~Hutter, ``Advanced skills by learning locomotion and local navigation end-to-end,'' in {\em 2022 IEEE/RSJ International Conference on Intelligent Robots and Systems (IROS)}, pp.~2497--2503, 2022.

\bibitem{petrović2018motion}
L.~Petrović, ``Motion planning in high-dimensional spaces,'' {\em arXiv preprint arXiv:1806.07457}, 2018.

\bibitem{app13148174}
K.~Almazrouei, I.~Kamel, and T.~Rabie, ``Dynamic obstacle avoidance and path planning through reinforcement learning,'' {\em Applied Sciences}, vol.~13, no.~14, 2023.

\bibitem{honerkamp2022learning}
D.~Honerkamp, T.~Welschehold, and A.~Valada, ``N$^2$m$^2$: Learning navigation for arbitrary mobile manipulation motions in unseen and dynamic environments,'' {\em IEEE Transactions on Robotics}, 2023.

\bibitem{mittal2023orbit}
M.~Mittal, C.~Yu, Q.~Yu, J.~Liu, N.~Rudin, D.~Hoeller, J.~L. Yuan, R.~Singh, Y.~Guo, H.~Mazhar, A.~Mandlekar, B.~Babich, G.~State, M.~Hutter, and A.~Garg, ``Orbit: A unified simulation framework for interactive robot learning environments,'' {\em IEEE Robotics and Automation Letters}, vol.~8, p.~3740–3747, June 2023.

\bibitem{miki2024confined}
T.~Miki, J.~Lee, L.~Wellhausen, and M.~Hutter, ``Learning to walk in confined spaces using 3d representation,'' in {\em 2024 International Conference on Robotics and Automation (ICRA)}, IEEE, 2024.

\bibitem{Frey_2022}
J.~Frey, D.~Hoeller, S.~Khattak, and M.~Hutter, ``Locomotion policy guided traversability learning using volumetric representations of complex environments,'' in {\em 2022 IEEE/RSJ International Conference on Intelligent Robots and Systems (IROS)}, IEEE, Oct. 2022.

\bibitem{miki2022elevation}
T.~Miki, L.~Wellhausen, R.~Grandia, F.~Jenelten, T.~Homberger, and M.~Hutter, ``Elevation mapping for locomotion and navigation using gpu,'' in {\em 2022 IEEE/RSJ International Conference on Intelligent Robots and Systems (IROS)}, pp.~2273--2280, IEEE, 2022.

\bibitem{ANYmal-D}
ANYbotics, ``Anymal-d datasheet.'' \url{https://www.anybotics.com/anymal-technical-specifications.pdf}.
\newblock Accessed: 2024-07-01.

\end{thebibliography}

\end{document}